%% file: acl_latex.tex
\documentclass[11pt]{article}

\usepackage[final]{acl}

\usepackage{times}
\usepackage{latexsym}

\usepackage[T1]{fontenc}
\usepackage[utf8]{inputenc}

\usepackage{microtype}

\usepackage{inconsolata}

\usepackage{amsmath}
\usepackage{amssymb}
\usepackage{mathtools}
\usepackage{amsthm}

\usepackage[capitalize,noabbrev]{cleveref}

\usepackage{times}
\usepackage{latexsym}
\usepackage{amsfonts}
\usepackage{bm}
\usepackage{multirow}
\usepackage{makecell}
\usepackage{tabulary}
\usepackage{algorithm}
\usepackage{algpseudocode}
\usepackage{graphicx}
\usepackage{booktabs}
\graphicspath{ {./figures/} }
\usepackage[textsize=tiny]{todonotes}

\input{math_commands}
\newcommand{\methodname}{\textsc{Resonance RoPE}}
\newcommand{\resyarn}{\textsc{Resonance YaRN}}
\newcommand{\datasetname}{\textsc{PosGen}}

\newcommand{\highfreqfeat}{pre-critical dimensions}
\newcommand{\lowfreqfeat}{post-critical dimensions}

\newcommand\blfootnote[1]{%
  \begingroup
  \renewcommand\thefootnote{}\footnote{#1}%
  \addtocounter{footnote}{-1}%
  \endgroup
}

\newtheorem{theorem}{Theorem}

\title{Resonance RoPE: Improving Context Length Generalization of\\ Large Language Models}

\author{
    Suyuchen Wang\textsuperscript{\rm 1,2},
    Ivan Kobyzev\textsuperscript{\rm 3},
    Peng Lu\textsuperscript{\rm 1},
    Mehdi Rezagholizadeh\textsuperscript{\rm 3}  \and
    Bang Liu\textsuperscript{\rm 1,2$\dagger$}
    \\
    \textsuperscript{\rm 1}DIRO, Universit\'e de Montr\'eal\quad \textsuperscript{\rm 2}Mila - Quebec AI Institute\quad \textsuperscript{\rm 3}Huawei Noah's Ark Lab\\
    \{\texttt{suyuchen.wang, peng.lu, bang.liu}\}\texttt{@umontreal.ca} \\
    \{\texttt{ivan.kobyzev, mehdi.rezagholizadeh}\}\texttt{@huawei.com}
}

\begin{document}
\maketitle

\begin{abstract}
This paper addresses the challenge of train-short-test-long (TSTL) scenarios in Large Language Models (LLMs) equipped with Rotary Position Embedding (RoPE), where models pre-trained on shorter sequences face difficulty with out-of-distribution (OOD) token positions in longer sequences. We introduce~\methodname, a novel approach designed to narrow the generalization gap in TSTL scenarios by refining the interpolation of RoPE features for OOD positions, significantly improving the model performance without additional online computational costs. Furthermore, we present~\datasetname, a new synthetic benchmark specifically designed for fine-grained behavior analysis in TSTL scenarios, aiming to isolate the constantly increasing difficulty of token generation on long contexts from the challenges of recognizing new token positions. Our experiments on synthetic tasks show that after applying~\methodname, Transformers recognize OOD position better and more robustly. Our extensive LLM experiments also show superior performance after applying~\methodname~to the current state-of-the-art RoPE scaling method, YaRN, on both upstream language modeling tasks and a variety of downstream long-text applications.\footnote{\url{https://github.com/sheryc/resonance_rope}.}
\end{abstract}

\section{Introduction}
\blfootnote{$^\dagger$Canada CIFAR AI Chair. Corresponding author.}
Recent advancements in Large Language Models (LLMs) have demonstrated their potential across a wide spectrum of natural language processing tasks, showcasing their ability to handle complex interactions, document analyses, professional writing, and advanced reasoning with a unified approach~\citep{openai_gpt-4_2023,touvron_llama_2023,touvron_llama_2023-1,jiang_mixtral_2024}. As these models are increasingly adapted for complex applications, challenges arise in scenarios requiring the comprehension or generation of long texts. Specifically, the train-short-test-long (TSTL) scenario~\citep{press_train_2022} highlights a limitation where LLMs, pre-trained on shorter sequences, struggle with out-of-distribution (OOD) token positions in longer sequences, impacting their performance in real-world applications~\citep{zhao_length_2023}.

Recent efforts to enhance TSTL performance have focused on LLMs equipped with Rotary Position Embedding (RoPE)~\citep{su_roformer_2023}, such as LLaMA~\citep{touvron_llama_2023,touvron_llama_2023-1} and Mistral~\citep{jiang_mistral_2023}, owing to their exceptional capabilities and widespread adoption. These initiatives aim to refine the test-time computation of RoPE position embedding by introducing a scaling factor to either the position index of each token~\citep{chen_extending_2023} or RoPE's base value~\citep{xiong_effective_2023,liu_scaling_2023,peng_yarn_2023}. These methods ensure that the position embeddings for out-of-distribution (OOD) positions remain within the range experienced during pre-training. This minimizes the need for the model to adapt to new position embedding value ranges, a task that is inherently difficult.

In this paper, we introduce~\methodname, a novel technique designed to further narrow the generalization gap on position embeddings in TSTL scenarios. Recognizing that RoPE's position embedding is governed by a complex, non-linear function, we posit that minimizing extrapolation on OOD positions, while crucial, is insufficient. We argue that \textit{it is equally vital to address the \textbf{interpolation} of RoPE features at the OOD positions.} By implementing~\methodname, we slightly scale each RoPE feature to correspond to an integer wavelength. This adjustment aligns each RoPE feature's wavelength with a specific token span length, enabling it to "resonate" with a particular local context length. This simple modification effectively reduces the generalization gap for over half of the position embedding features in LLaMA and LLaMA2 under TSTL scenarios. Furthermore, our approach is compatible with RoPE and any RoPE-based scaling techniques, enhancing their performance in TSTL situations without the need for additional computational resources during training or inference.

Additionally, to facilitate further research on position embeddings, we present a new synthetic benchmark tailored for TSTL scenarios, named~\datasetname. Improving position embeddings for TSTL requires a detailed analysis of the cause of failures in handling longer contexts. However, current benchmarks, such as those measuring perplexity in long context~\citep{rae_compressive_2019,huang_efficient_2021,wu_memorizing_2021} and most synthetic TSTL tasks~\citep{liu_transformers_2023,kazemnejad_impact_2023} face a common issue: the difficulty of generating the next token increases with context length. This makes it difficult to determine whether a model's failure is due to its inability to generate more complex tokens or its failure to recognize out-of-distribution (OOD) positions. \datasetname~addresses this limitation by standardizing the difficulty level of token generation across all positions. This ensures that any observed shortcomings are directly related to the model's inability to identify and handle new token positions effectively.

Our contributions in this study are threefold:
\begin{enumerate}
    \item We propose~\methodname, an innovative modification to RoPE based on an in-depth analysis of the wavelengths of RoPE features, aiming to narrow the generalization gap in TSTL scenarios across RoPE and similar RoPE-based scaling techniques, without necessitating extra computational resources during runtime.
    \item We present~\datasetname, a newly developed synthetic benchmark tailored for TSTL scenarios. This benchmark is specifically designed to disentangle the complexities associated with generating tokens in longer contexts from the challenges posed by recognizing new positions or position embedding values.
    \item Through rigorous testing of~\methodname~on both RoPE and YaRN within the~\datasetname~benchmark, we demonstrate its ability to enhance performance on out-of-distribution (OOD) positions, surpassing existing methods that do not include~\methodname. Moreover, when applied to YaRN,~\methodname~further improves LLM's length extrapolation ability, as evidenced by lower perplexity in upstream TSTL language modeling and enhanced outcomes in downstream tasks involving lengthy contexts. % Emphasize LLM in the second part.
\end{enumerate}

\section{Related Work}

\subsection{Scaling of RoPE Position Encoding}

Recent efforts in extending LLMs' context window focus on manipulating position embedding (PE), particularly RoPE~\citep{su_roformer_2023}, which is used in LLMs like LLaMA~\citep{touvron_llama_2023, touvron_llama_2023-1} and Mistral~\citep{jiang_mistral_2023}. Main strategies include embedding scaling~\citep{chen_extending_2023, liu_scaling_2023, peng_yarn_2023} and randomizing token positions~\citep{ruoss_randomized_2023, zhu_pose_2023}. Our emphasis is on the embedding scaling strategies.

Existing embedding scaling strategies adjust position embedding for longer sequences to match the pre-training range, avoiding feature extrapolation. For instance, \citet{chen_extending_2023} compresses position indices to fit the pre-training range, extending LLaMA's~\citep{touvron_llama_2023} context to 16K with 1,000 steps of fine-tuning. Alternatively, \citet{liu_scaling_2023, roziere_code_2023, xiong_effective_2023} modify RoPE's rotary base and employ fine-tuning on extended sequences, termed Adjusted Base Frequency (ABF) or "NTK-aware" scaling. Code LLaMA~\citep{roziere_code_2023} achieved 16K context length with this method after 10,000 fine-tuning steps. YaRN~\citep{peng_yarn_2023} improved NTK-aware scaling by segmenting RoPE features and applying tailored extrapolation strategies, achieving 64K context length for LLaMA2~\citep{touvron_llama_2023-1} with 400 fine-tuning steps. Distinguishingly, our~\methodname~focus on reducing feature interpolation on OOD positions, which we argue is another important factor in improving the length extrapolation capability of Transformer. 

\subsection{Long Context Evaluations}

Evaluations of Transformer-based LLMs' long-context capabilities are twofold: synthetic task assessments for length extrapolation strategies and real-world task evaluations at the LLM scale. Synthetic evaluations target simple tasks such as long sequence classification~\cite{tay_long_2020} and arithmetic language modeling~\citep{liu_transformers_2023, kazemnejad_impact_2023}.
LLM scale evaluations measure metrics such as perplexity (PPL) in extensive text corpora (e.g., PG19~\citep{rae_compressive_2019}, GovReport~\citep{huang_efficient_2021}, GitHub~\citep{wu_memorizing_2021}) and complex tasks including summarization, question answering, and mathematical reasoning~\citep{an_l-eval_2023, bai_longbench_2023, shaham_zeroscrolls_2023}.

\section{Background}

\subsection{Rotary Position Embedding (RoPE)}

In Transformers~\citep{vaswani_attention_2017}, the self-attention scores are softmax-normalized scaled attention logits $\vq^\top \vk$:
\begin{equation*}
    a_{m,n}=\text{Softmax}\left(\frac{{\vq_m}^\top{\vk_n}}{\sqrt{d}}\right)
\end{equation*}

Suppose the input to a single attention head is $\vx_1, \vx_2, \ldots, \vx_l\in \R^d$, where $l$ is the sequence length and $d$ is the dimension of an attention head. RoPE injects the position information of each token into the $\vq$ and $\vk$ vectors by the following equations in the complex space:
\begin{align}
    \vq_{m, [2j:2j+1]}&=\mW_q\vx_m e^{im\theta_j} \notag \\
    \vk_{m, [2j:2j+1]}&=\mW_k\vx_m e^{im\theta_j} \notag \\
    \theta_j&=b^{\frac{-2j}{d}}, \label{eqn:theta}
\end{align}
where $\mW_q, \mW_k$ are trainable parameters, and $b$ is a constant called the rotary base, which is set to $10,000$ ~\citep{su_roformer_2023} or other integers or fractions~\citep{xiong_effective_2023,peng_yarn_2023}. This form makes the dot product between the $m$-th query $\vq_m$ and $n$-th key $\vk_n$ only depend on the input $\vx_m, \vx_n$ and their relative distance $(m-n)$:
\begin{align*}
    &\langle\vq_{m, [2j:2j+1]}, \vk_{n, [2j:2j+1]}\rangle\\
    =&\Re\left[\vq^*_{m, [2j:2j+1]}\vk_{n, [2j:2j+1]}\right]\\
    =&\Re\left[\left(\mW_q \vx_m\right)^*\left(\mW_k \vx_n\right) e^{i(m-n)\theta_j}\right]\\
    =&g(\vx_m, \vx_n, m-n) .
\end{align*}
RoPE's real-number implementation divides the $d$-dimension space into multiple $2$-dimensional subspaces and applies real rotation matrix to each of them. Formally, define 
% $\mR_{\Theta}^d\in\R^{d\times d}$:
a $d\times d$ block-diagonal matrix:
\begin{equation}
    \mR^d_{\Theta,m}=
	\begin{pmatrix}
        \tiny
		\mR_{\theta_0,m} & \cdots & \cdots & \mathbf{0} \\
        \mathbf{0} & \mR_{\theta_1,m} & \cdots & \mathbf{0} \\
        \vdots & \vdots & \ddots & \vdots \\
        \mathbf{0} & \mathbf{0} & \cdots & \mR_{\theta_{\frac{d}{2}-1},m} \\
	\end{pmatrix} ,
    \label{eqn:rope-feature}
\end{equation}
where $\Theta=\{\theta_0, \theta_1, \cdots, \theta_{\frac{d}{2}-1}\}$, and each $\mR_{\theta_j,m}$ is a $2\times 2$ rotation matrix:
\begin{equation}
    \mR_{\theta_j,m} = 
	\begin{pmatrix}
        \tiny
		\cos{m\theta_j}&-\sin{m\theta_j}\\
		\sin{m\theta_j}&\cos{m\theta_j} \\
	\end{pmatrix} .
    \label{eqn:rope-subfeature}
\end{equation}
RoPE computes the attention logit $\vq^\top \vk$ as follows:
\begin{align}
    \vq_m&=\mR^d_{\Theta,m}\mW_q\vx_m \label{eqn:q}\\
    \vk_n&=\mR^d_{\Theta,n}\mW_k\vx_n \label{eqn:k}\\
    \vq_m^\top \vk_n&=\vx_m^\top \mW_q \mR^d_{\Theta,n-m} \mW_k \vx_n \label{eqn:qk}
\end{align}
For each two dimensions $[2j:2j+1]$ of $\vq$ and $\vk$, its corresponding $\theta_j$ reflects a temporal wavelength $\lambda_j$. This wavelength describes the token length for the corresponding RoPE features to encounter approximately the same rotary angle $m\theta_j$ in Equation~\ref{eqn:rope-subfeature}:
\begin{equation}
    \label{eqn:wavelength}
    \lambda_j=\frac{2\pi}{\theta_j}=2\pi b^{\frac{2j}{d}}
\end{equation}
 As an example, the wavelengths of LLaMA / LLaMA2's RoPE features range from $2\pi\approx 6.28$ for $\theta_0$ to $2*10000^{126/128}\pi\approx 54410.14$ for $\theta_{\frac{d}{2}-1}$.

\subsection{Critical Dimensions of RoPE}

In a TSTL scenario~\citep{press_train_2022}, one takes a model trained on texts with lengths up to $L$, and tests it on a task with input lengths up to $L'=sL$, with the scaling factor $s>1$.
Recently, \citet{liu_scaling_2023} discovered that there may exist two ``critical dimensions'' in RoPE features, which correspond to the dimensions $[2c:2c+1]$ that satisfies $\lambda_c \ge L$ and $\lambda_{c-1} < L$. The dimensions of RoPE features above and below the critical dimension (which we denote as ``\lowfreqfeat'' and ``\highfreqfeat'', respectively) have different behaviors in TSTL: for~\lowfreqfeat~(i.e., $j>c$), since their wavelengths satisfy $\lambda_j>L$, the training corpus does not cover all possible rotary angles $m\theta_j$ on a unit circle. Thus, these dimensions will encounter OOD value range on longer sequences. This is not an issue for~\highfreqfeat~due to their shorter temporal wavelengths.
 
The concept of RoPE's critical dimensions implicitly guides the development of RoPE scaling methods. For example, previous RoPE scaling methods~\citep{chen_extending_2023,xiong_effective_2023,peng_yarn_2023} mainly focus on reducing or avoiding value extrapolation on~\lowfreqfeat, and minimize post-training modifications to the~\highfreqfeat.

\subsection{Yet another RoPE extensioN (YaRN)}
\label{sec:yarn}

YaRN~\citep{peng_yarn_2023} is the current state-of-the-art RoPE scaling method for TSTL. It introduces the ``NTK-by-parts'' scaling for RoPE, which applies different scaling strategies to each RoPE feature according to its temporal wavelength.

In a TSTL scenario with scaling factor $s$, YaRN scales the wavelength of the $j$-th RoPE feature $\lambda_j$ to $\hat{\lambda_j}$ and further fine-tune the model:
\begin{equation*}
    \hat{\lambda_j}=(1-\gamma_j)s\lambda_j+\gamma_j\lambda_j,
\end{equation*}
where $\gamma_j$ is a piece-wise function depending on its corresponding wavelength $\lambda_j$, and two hyperparameters $\alpha$ and $\beta$:
\begin{equation*}
    \gamma_j=\left\{  
     \begin{aligned}
     &1, &\text{if}~\lambda_j<L/\beta \\
     &0, &\text{if}~\lambda_j>L/\alpha \\
     &\frac{L/\lambda_j-\alpha}{\beta-\alpha}, &\text{otherwise}
     \end{aligned}
\right.  
\end{equation*}

Empirically, for the LLaMA family, \citet{peng_yarn_2023} suggests using $\alpha=1$ and $\beta=32$. This setting avoids value range extrapolation on~\lowfreqfeat, while reducing modifications to the original~\highfreqfeat.

In addition to the ``NTK-by-parts'' RoPE scaling strategy mentioned above, YaRN also comprises a scaling strategy on the attention scores, which reduces the change in the entropy of the attention score on longer sequences. We maintain the complete design of YaRN in our experiments, but our analysis will focus on its RoPE scaling strategy.

\section{Proposed Method:~\methodname}

In this section, we introduce~\methodname, a universal improvement for RoPE and RoPE-based scaling methods to (further) improve their length extrapolation performance.

Suppose we abstract RoPE's Equation~\ref{eqn:q},~\ref{eqn:k}: for any $\vx \in \mathbb{R}^d$, we define $f(\vx,m)=\mR^d_{\Theta,m}\mW \vx$. In a TSTL scenario where we generalize an LLM from length $L$ to length $L'$, let us denote a scaled RoPE function by $\tilde{f}$. To perform well on OOD positions it should reduce the \textit{feature gap} $h(\tilde{f})$ between token features seen during training and token features after scaling that we can define for each  $i$-th feature as:
\begin{equation}
\label{eq:gap}
h_i(\tilde{f})=\max_{\vx\in\mathbb{X}}\min_{\begin{subarray}{c}
    m\in\{0,\cdots,L-1\}
    \\n\in\{L,\cdots,L'-1\}
    \end{subarray}} |\tilde{f}(\vx,m)_i -\tilde{f}(\vx,n)_i |,
\end{equation}
where $i = 0, \dots, d-1$ and $\mathbb{X} \subset \mathbb{R}^d$ is the set of feature vectors to which we apply a position embedding. Note that the formulation of the feature gap is similar to the ``embedded vector distance'' metric proposed by \citet{xiong_effective_2023}. However, these two metrics target totally different aspects of RoPE scaling methods. A more detailed comparison can be found in Appendix~\ref{appx:comparison}.

Existing RoPE scaling methods~\citep{xiong_effective_2023,peng_yarn_2023} mainly focus on the~\lowfreqfeat~ of RoPE, since the rotary angle $m\theta_j$ on these dimensions extrapolates on OOD positions, hence creating a feature gap. In this section, we argue that reducing RoPE's feature interpolation on the~\highfreqfeat~ is also beneficial for better length extrapolation.

\begin{figure}[t!]
  \includegraphics[width=\linewidth]{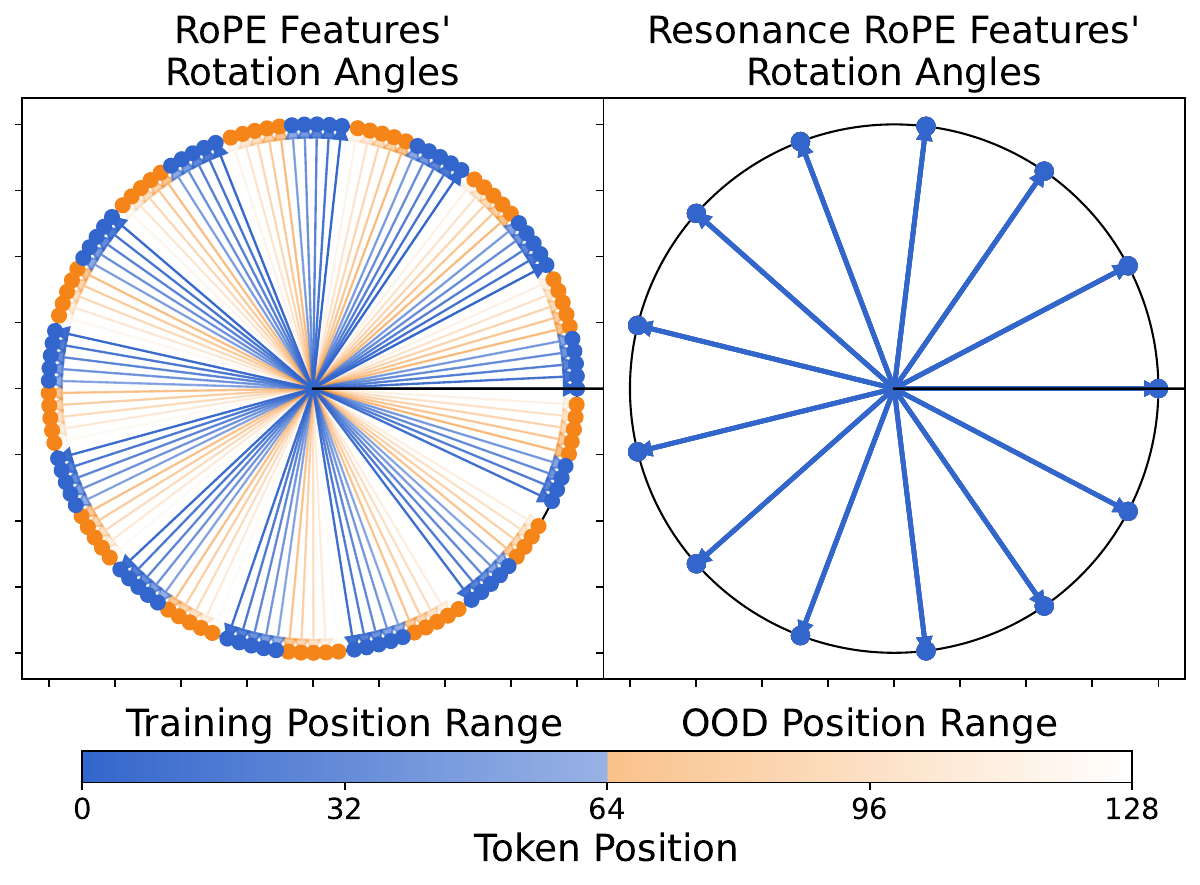}
  \caption{An illustration of RoPE's rotation angles $m\theta_6$ and~\methodname's rotation angles $m\tilde{\theta}_6$ in Eqn.~\ref{eqn:rope-subfeature} in a TSTL scenario with training max length $64$ and testing max length $128$. RoPE's non-integer feature wavelengths create a feature gap between the RoPE features of the training and OOD testing positions, while~\methodname~reduces this gap to 0.}
  \vspace{-5mm}
  \label{fig:explanation}
\end{figure}

Due to a non-linear relationship between RoPE feature $\mR^\Theta_m$ and the token position $m$ in Equation~\ref{eqn:rope-subfeature}, the interpolation on RoPE features is potentially hard for the model to generalize to.
We found that such potentially hard interpolation appears on the~\highfreqfeat~$[0:2c-1]$, which have wavelengths $\lambda_j$ shorter than the pre-trained sequence length $L$. By default, the rotary base $b$ of RoPE features is an integer or a fraction, which makes their wavelength $\lambda_j=2\pi b^{\frac{2j}{d}}$ not an integer. As the position index $m\in\sN$ increases, a phase shift of $\Delta\phi$ occurs for the rotary angle $m\theta_j$ after each full rotation. This could potentially result in a large distribution gap between the RoPE features on positions seen during training and the OOD positions. This phenomenon is illustrated in Figure~\ref{fig:explanation}.

\begin{algorithm}[t!]
\caption{Pseudocode of~\methodname.}
    \label{alg:resonance_rope}
    \begin{algorithmic}
        \Require $\theta_0, \theta_1, \cdots, \theta_{\frac{d}{2}-1}\in \Theta$
        \For{$i\in \{0,1,\cdots,\frac{d}{2}-1\}$}
            \State $\lambda_i=2\pi/\theta_i$
            \State $\tilde{\lambda}_i=\text{round}(\lambda_i)$ \Comment{{\small Round to integer wavelength}}
            \State $\tilde{\theta}_i=2\pi/\tilde{\lambda}_i$
        \EndFor
        \State $\tilde{\Theta}=\{\tilde{\theta}_0, \tilde{\theta}_1, \cdots, \tilde{\theta}_{\frac{d}{2}-1}\}$
        \State Compute $\mR^d_{\tilde{\Theta}}$ by Equation~\ref{eqn:rope-feature}
        \State Compute $\vq$, $\vk$ by Equation~\ref{eqn:q},~\ref{eqn:k}
    \end{algorithmic}
\end{algorithm}

We tackle this issue by developing a synergistic modification to the conventional RoPE embedding, referred to as~\methodname. It aims to identify the optimal angular frequency that minimizes the interpolation gap, which ensures the corresponding wavelength closely matches the original one while imposing alignment of the wavelength to an integer.
More specifically, for a given angular frequency set of RoPE $\Theta=\left\{\theta_1, \theta_2, \ldots, \theta_{d/2}\right\}$, we round their wavelengths to their nearest integer to eliminate new rotary angles on each feature. We provide a pseudocode for~\methodname~in Algorithm~\ref{alg:resonance_rope}. 

After applying this technique, each RoPE feature repeats after $\tilde{\lambda}_i$ tokens, and therefore ``resonates'' with a specific span length and eliminates the interpolation gap between pre-trained and OOD positions on~\highfreqfeat. We illustrate the effect of~\methodname~on RoPE's feature gap on one of the~\highfreqfeat~in Figure~\ref{fig:explanation}. Moreover, we can prove the feature gap reducing ability of our method. As for above, we formalize~\methodname's computation rule as $\tilde{f}(\vx,m)=\mR^d_{\tilde{\Theta},m}\mW \vx$.

\begin{theorem}
\label{thm:generalization}
For a RoPE-equipped model with context window $L$,~\methodname~$\tilde{f}$ reduces the feature gap on~\highfreqfeat~to $0$. Specifically, $\forall \vx\in\sX$, $\forall n\in\sN\backslash\{0,\cdots,L-1\}$, we have:
\begin{equation*}
% \max_{\vx\in\mathbb{X}}
\min_{m \in \{0,\cdots,L-1\}
    } | \tilde{f}(\vx,m)_i - \tilde{f}(\vx,n)_i|=0
\end{equation*}
for all $i = 0,\dots,2c-1$.
\end{theorem}
See the proof in Appendix~\ref{appx:proof}.
Note that although each pre-critical RoPE feature $\mR_{\tilde{\theta}_j, m}$ repeats, the combination of all
% ~\highfreqfeat~$\mR_{\{\tilde{\theta}_j,j<c\}}^m$ 
$\{\mR_{\tilde{\theta}_j, m} \}_{j<c}$
only repeats after the least common multiple (LCM) of all~\highfreqfeat's wavelengths. For LLaMA2, this LCM value is greater than $7\times 10^{51}$.

Because of its simplicity,~\methodname~can be applied on top of RoPE and all RoPE-based scaling methods to reduce their feature gap in TSTL and further improve their performance. Meanwhile, this method only involves an offline computation of the scaled $\theta$, thus introducing no online computation overhead. 

\section{Evaluating Position Embeddings with~\datasetname}

\begin{figure*}[t!]
  \includegraphics[width=\textwidth]{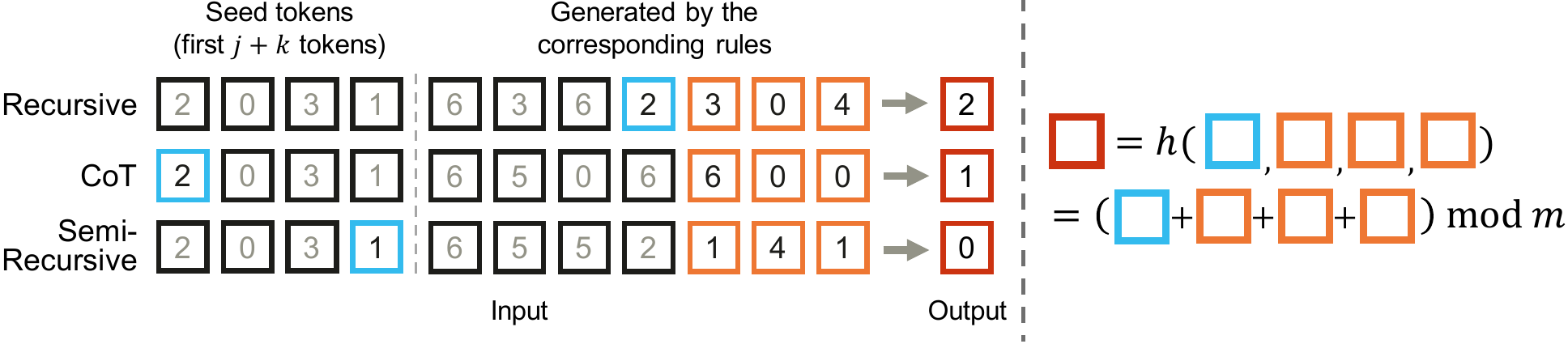}
  \caption{An example of the three subtasks of~\datasetname. This figure shows the process of generating the $12$th token shown in the red boxes for each subtask. In this example, $h$ is a modular addition task with the modulus $m=7$ and the difficulty-controlling parameters $j=1, k=3$. The output token depends on: (1) only the local $j+k$ tokens in the recursive task; (2) $k$ local tokens and the beginning $j$ tokens in the CoT task; and (3) $k$ local tokens and $j$ tokens with a varied dependency distance in the semi-recursive task.}
  \label{fig:dataset}
  \vspace{-5mm}
\end{figure*}

In this section, we propose our new position embedding evaluation suite:~\datasetname, based on an analysis of common failure patterns on existing position embedding evaluation methods.

We consider a next token prediction task, where we expect the model to generate the token $x_l$ given the input sequence $\{x_0,\cdots,x_{l-1}\}$. In TSTL scenarios, when a model succeeds in correctly generating a token up to position $L$ but fails systematically afterwards, we observe two failure patterns:

\begin{itemize}
    \item \textbf{Failure due to harder algorithmic difficulty on generating later tokens.}
    The rule of generating a new token $x_l$ may vary with the sequence length $l$.
    Generally, tokens placed later in the sequence depend on more context tokens, which incurs a more complex dependency pattern. During training on shorter sequences, the model only learns the token dependency rules involving up to $L$ tokens, and might fail on longer sequences because it has never been exposed to the more complex dependency rules.
    \item \textbf{Failure due to unrecognized new token positions.} The difference between training and testing lengths in the TSTL setting creates a feature gap between the position indices or position embeddings in training and inference. This feature gap makes it difficult for the model to generalize to new positions due to unrecognized features. RoPE scaling methods mainly focus on reducing this type of length extrapolation failure.
\end{itemize}

Currently, neither perplexity-based evaluations~\citep{rae_compressive_2019, huang_efficient_2021, wu_memorizing_2021} nor synthetic TSTL evaluations~\citep{kazemnejad_impact_2023,liu_transformers_2023} can effectively distinguish these two failure patterns, since the token generation difficulty tends to increase with respect to the sequence length in these tasks. To facilitate research on better position representations, we design~\datasetname, which controls the difficulty in generating tokens throughout the sequence to be identical, which effectively distinguishes the two types of TSTL failures. Failures in this benchmark are only due to the inability to recognize new token positions in TSTL scenarios.

Our~\datasetname~framework comprises three sub-tasks, with each extracting the general token dependency pattern of a different type of reasoning task. Suppose that we define a fixed function $h:\mathbb{V}^{j+k}\to\mathbb{V}$, where $\mathbb{V}$ is the model's vocabulary and $j,k$ are predefined constants controlling the task's difficulty. The three subtasks of~\datasetname~are as follows:
\begin{enumerate}
    \item \textbf{Recursive.} This task simulates the token dependency pattern of generating a Fibonacci-style sequence, where new tokens depend on $j+k$ neighboring tokens only: $x_l=h(x_{l-(j+k))},\cdots,x_{l-1})$ when $l\ge j+k$.
    \item \textbf{Chain-of-Thought (CoT).} This task simulates the token dependency pattern of CoT reasoning~\citep{wei_chain_2022}, where new tokens depend on $k$ neighboring tokens (simulating the previous reasoning step) and $j$ tokens in the front (simulating the original question): $x_l=h(x_0,\cdots,x_{j-1},x_{l-k},\cdots,x_{l-1})$ when $l\ge j+k$.
    \item \textbf{Semi-recursive.} This task simulates the token dependency pattern of the last-letter concatenation task~\citep{zhou_least--most_2022}, where new tokens depend on both $k$ neighboring tokens (simulating the current progress) and $j$ tokens with varied distances according to a specific rule (simulating the word sequence): $x_l=h(x_{\floor{l-(j+k)/2}-j},\cdots,x_{\floor{l-(j+k)/2}-1},\\x_{l-k},\cdots,x_{l-1})$ when $l\ge j+k$.
\end{enumerate}

\label{sec:ood_accuracy}
Based on the equation for each subtask, when given the first $j+k$ tokens, one can generate a sequence with unlimited length as the ground truth sequence. We show an example of~\datasetname~in Figure~\ref{fig:dataset}. As a TSTL benchmark, we train a model on a subtask with sequence length up to $L$, and evaluate the model's accuracy on a longer sequence with length $L'>L$ generated by the same rule on the unseen positions $L<m\le L'$, which we refer to as the ``OOD Accuracy'' (OOD Acc).
This metric measures how well a model can recognize the OOD positions and continue following the generation rule learned during training.
% In this configuration, the only failure pattern with a low OOD Accuracy would only be due to the unrecognized new token positions in a longer sequence.
% This is because the algorithmic difficulty in generating each token is controlled by $h$ and the constants $j,k$, thus does not change with the sequence length.
As a benchmark for position embeddings, a standard usage of this benchmark is to train a small Transformer (e.g., a 2-layer Transformer as used in our experiments) with different position embeddings on its training set with only short sequences, and test its OOD Accuracy on the test set with longer sequences.
We provide our experiment setting for~\datasetname~in more details in Section~\ref{sec:synthetic_experiments} and Appendix~\ref{appx:synthetic_experiments}.

\section{Experiments}

We evaluate~\methodname~on three different TSTL tasks: a small-scale evaluation on our proposed~\datasetname~task, and LLM-scale evaluations with LLaMA2-Chat~\citep{touvron_llama_2023-1} on both language modeling perplexity and real-world long context applications.

\subsection{Synthetic Task Evaluation}

\begin{figure*}[t!]
    \centering
    \includegraphics[width=0.98\textwidth]{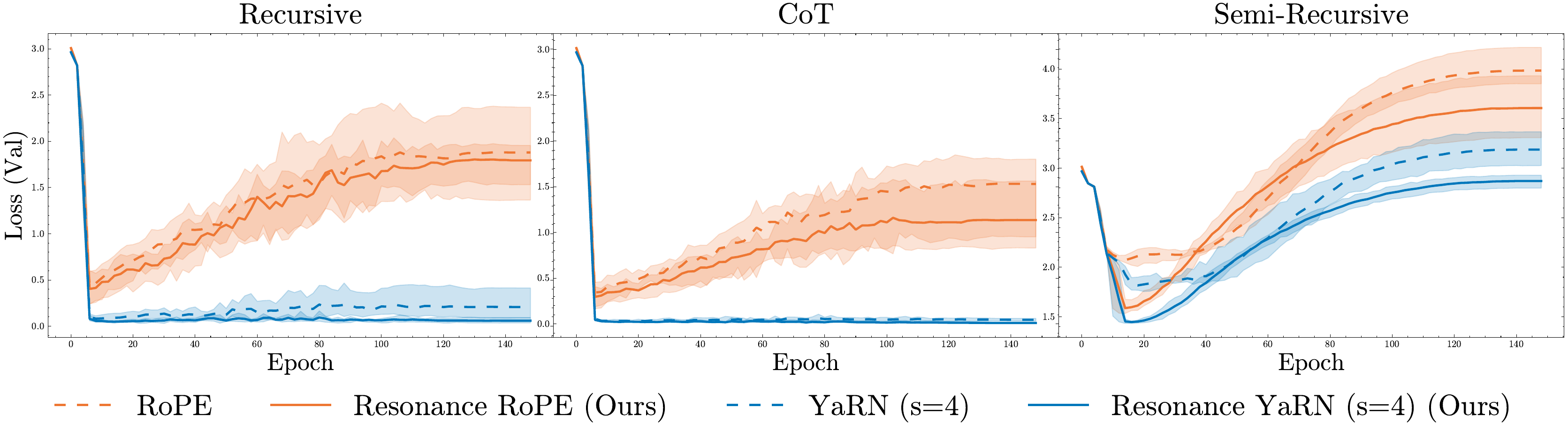}
    \caption{The validation loss curves of Transformers using RoPE and YaRN PEs with and without our \textsc{Resonance} scaling on the three subtasks of~\datasetname.}
  \label{fig:synthetic_val_loss}
  \vspace{-3mm}
\end{figure*}

\subsubsection{Experiment Setup}
\label{sec:synthetic_experiments}
We first apply~\methodname~on RoPE and YaRN, assessing the model's performance on~\datasetname~for unseen position recognition. We test on a  modular addition task, which was proved to be learnable by a one-layer Transformer~\citep{nanda_progress_2023}. We configured $j=1, k=3$, and defined $h(x_0, x_1, x_2, x_3)=\sum_{i=0}^3{x_i}\mod 17$ with vocabulary $\mathbb{V}=\{0,\ldots,16\}$.

Our experiments involved training a two-layer Transformer. Each layer follows T5-Small's configurations~\citep{raffel_exploring_2023} except for the position embeddings. In this model, each attention head has $64$ dimensions. We apply different RoPE-based embeddings with the rotary base equal to $10,000$. The models are trained on sequences of length $L=64$, and evaluating on lengths of $L'=256$ for OOD Accuracy. In this experiment setting, each head has $32$ RoPE features, out of which the first $17$ are pre-critical dimensions with a wavelength less than the maximum training length. We generated 10,000 training sequences, and 1,000 each for validation and testing, and ensured that the first $j+k=4$ tokens in each sequence do not overlap to testify whether the model learns the correct generation mechanism. We averaged results over $5$ seeds. A more detailed setting is provided in Appendix~\ref{appx:synthetic_experiments}.

\subsubsection{Results and Analysis}

\begin{table}[t]
\centering % used for centering table
\resizebox{\columnwidth}{!}{
\begin{tabular}{c|ccc}
\toprule
\multicolumn{1}{c}{Setting} & Recursive & CoT & Semi-Rec. \\
\midrule
\small{RoPE} & $\textbf{65.29}{\scriptstyle \pm \textbf{0.43}}$ & $69.56{\scriptstyle \pm 0.33}$ & $17.96{\scriptstyle \pm 0.03}$ \\
\small{Res. RoPE (Ours)} & $62.64{\scriptstyle \pm 0.15}$ & $\textbf{75.25}{\scriptstyle \pm \textbf{0.10}}$ & $\textbf{29.78}{\scriptstyle \pm \textbf{0.07}}$ \\
\midrule
\small{YaRN} & $95.93{\scriptstyle \pm 0.04}$ & $98.71{\scriptstyle \pm 0.00}$ & $33.70{\scriptstyle \pm 0.04}$ \\
\small{Res. YaRN (Ours)} & $\textbf{98.30}{\scriptstyle \pm \textbf{0.00}}$ & $\textbf{99.58}{\scriptstyle \pm \textbf{0.00}}$ & $\textbf{48.46}{\scriptstyle \pm \textbf{0.03}}$ \\
\bottomrule
\end{tabular}}
\caption{The accuracy on OOD Positions (OOD Acc.) on~\datasetname's test set. All results are in percentage (\%). We report both the mean and variance across five runs with different random seeds. We compare the same RoPE-based PE with or without our \textsc{Resonance} scaling. The best performance for each pair of settings on each subtask is marked in \textbf{Bold}.} % title of Table
\label{tab:synthetic} % is used to refer this table in the text
\vspace{-5mm}
\end{table}

Table~\ref{tab:synthetic} displays the comparison of the OOD accuracy. In most cases, \methodname~and~\resyarn~outperform their counterparts lacking the Resonance technique, showcasing significantly better performance and reduced variance in OOD scenarios. This improvement indicates a superior adaptation to OOD position embeddings through minimized Positional Encoding (PE) interpolation. An exception is observed when applying~\methodname~to the Recursive subtask, likely due to the dominance of extrapolated \lowfreqfeat~in OOD positions. This issue can be mitigated by employing a RoPE scaling technique such as YaRN, which effectively counters the extrapolation of \lowfreqfeat. Among all configurations, \resyarn~exhibits the highest OOD performance, demonstrating the synergy between RoPE scaling methods and the Resonance technique.

Figure~\ref{fig:synthetic_val_loss} plots validation losses against training epochs for different PEs, illustrating the training dynamics. The introduction of the Resonance technique leads to a reduction in the lowest validation loss for both RoPE and YaRN, with~\methodname~achieving even lower validation losses than YaRN in the Semi-Recursive subtask. Furthermore, the validation loss trajectories for \methodname~and \resyarn~remain lower than those of their counterparts in all subtasks, further demonstrating the enhanced OOD generalization capability of our approach.

\begin{table*}[t]
\centering % used for centering table
\resizebox{\textwidth}{!}{
\begin{tabular}{rc|cccccc|c}
\hline
\multicolumn{1}{c}{Setting} & Ctx Len. & Coursera & GSM & QuALITY & TOEFL & CodeU & SFiction & Avg.  \\
\hline
\multicolumn{9}{c}{\textbf{\small LLaMA2-Chat 7B}} \\
\hline
 Dynamic NTK-Aware (no FT)           & 32K                 & 31.98    & \textbf{32.00}  & 34.65   & \textbf{59.11} & 1.11  & 36.72    & 32.59 \\
                             NTK-Aware ($s=8$, no FT)            & 32K                 & \textbf{36.77}    & 3.00   & 26.73   & 34.2  & 1.11  & 50.78    & 25.43 \\ 
                             \hline
                             YaRN ($s=8$, FT@$32$K, 
 $50$ epcs.)               & 32K                 & 36.05    & 19.00  & 33.17   & 50.56 & \textbf{4.44}  & 56.25    & 33.24 \\
                             Resonance YaRN ($s=8$, FT@$32$K, $50$ epcs.)     & 32K                 & \underline{36.48}    & 22.00  & 34.16   & 55.76 & 0.00  & \underline{57.03}    & 34.24 \\
                            \hline
                             YaRN ($s=8$, FT@$4$K, $400$ epcs.)                & 32K                 & 35.03    & 24.00  & \underline{37.62}   & \underline{57.62} & \textbf{4.44}  & 60.94    & \underline{36.61} \\
                             Resonance YaRN ($s=8$, FT@$4$K, $400$ epcs.)      & 32K                 & 36.34    & \underline{27.00}  & \textbf{40.59}   & 56.51 & \underline{3.33}  & \textbf{61.72}    & \textbf{37.58} \\
\hline
\multicolumn{9}{c}{\textbf{\small LLaMA2-Chat 13B}} \\
\hline
 Dynamic NTK-Aware (no FT)           & 16K                 & 29.22    & \textbf{39.00}  & 40.59   & 63.94 & 1.11  & 39.84    & 35.62 \\
                             NTK-Aware ($s=4$, no FT)            & 16K                 & 40.26    & 21.00  & 38.12   & 65.43 & 1.11  & 46.88    & 35.47 \\  
                            \hline
                             YaRN ($s=4$, FT@$16$K, $100$ epcs.)               & 16K                 & 38.08    & \textbf{39.00}  & \underline{43.07}   & 65.43 & 0.00  & \textbf{63.28}    & \underline{41.48} \\
                             Resonance YaRN ($s=4$, FT@$16$K, $100$ epcs.)     & 16K                 & 38.66    & \textbf{39.00}  & \textbf{43.56}   & 65.06 & 1.11  & \underline{62.50}    & \textbf{41.65} \\
                            \hline
                             YaRN ($s=4$, FT@$4$K, $400$ epcs.)                & 16K                 & \underline{41.72}    & 34.00  & 41.09   & \textbf{66.91} & \underline{2.22}  & 48.44  & 39.06 \\
                             Resonance YaRN ($s=4$, FT@$4$K, $400$ epcs.)      & 16K                 & \textbf{41.86}    & \underline{35.00}  & 42.57   & \underline{65.80} & \textbf{5.56}  & 48.44  & 39.87 \\
\hline
\end{tabular}}
\caption{Long text evaluations on some closed-ended tasks in L-Eval. ``Ctx Len'' means the target context length of the model after scaling its PE. ``FT@$32$K, $50$ epcs'' means the model is fine-tuned on $32$K sequence length for $50$ epochs. The settings with ``no FT'' are not fine-tuned after modifying its position embedding. We highlight the best and second-best performance for each base model in \textbf{Bold} and \underline{Underline}, respectively.} % title of Table
\label{tab:leval} % is used to refer this table in the text
\vspace{-5mm}
\end{table*}

\subsection{LLM Fine-tuning Evaluation}
\label{sec:llm_experiments}

\subsubsection{Experiment Setup}

In this section, we apply our proposed~\methodname~to the current state-of-the-art RoPE scaling method, YaRN~\citep{peng_yarn_2023}.
More specifically, we replace the original position embeddings of LLaMA2 7B and 13B~\citep{touvron_llama_2023-1} with a series of scaled position embeddings, including the NTK-Aware scaling~\citep{bloc97_ntk-aware_2023,xiong_effective_2023,liu_scaling_2023}, Dynamic NTK-Aware Scaling~\citep{peng_yarn_2023,roziere_code_2023}, and YaRN~\citep{peng_yarn_2023}.

For YaRN and~\resyarn, We use a scaling factor of $8$ and $4$ for LLaMA2 7B and 13B to extend their context window from $4$K to $32$K and $16$K, respectively. 
For the configurations that require fine-tuning, we fine-tune the LLM with the scaled position embedding on the training set of PG19~\citep{rae_compressive_2019} with the fine-tuning setting and hyperparameters adopted directly from YaRN~\citep{peng_yarn_2023}, with the only difference being that we control the total training token count to be approximately $100$M. A more detailed fine-tuning setting can be found in Appendix~\ref{appx:llm_setting}. We test the model's performance on two TSTL scenarios: language modeling evaluation on long-text sequences and long-text downstream application performance.

\subsubsection{Perplexity on Long Sequence}
\label{sec:perplexity_experiments}
\begin{figure}[t]
    \centering
    \includegraphics[width=\columnwidth]{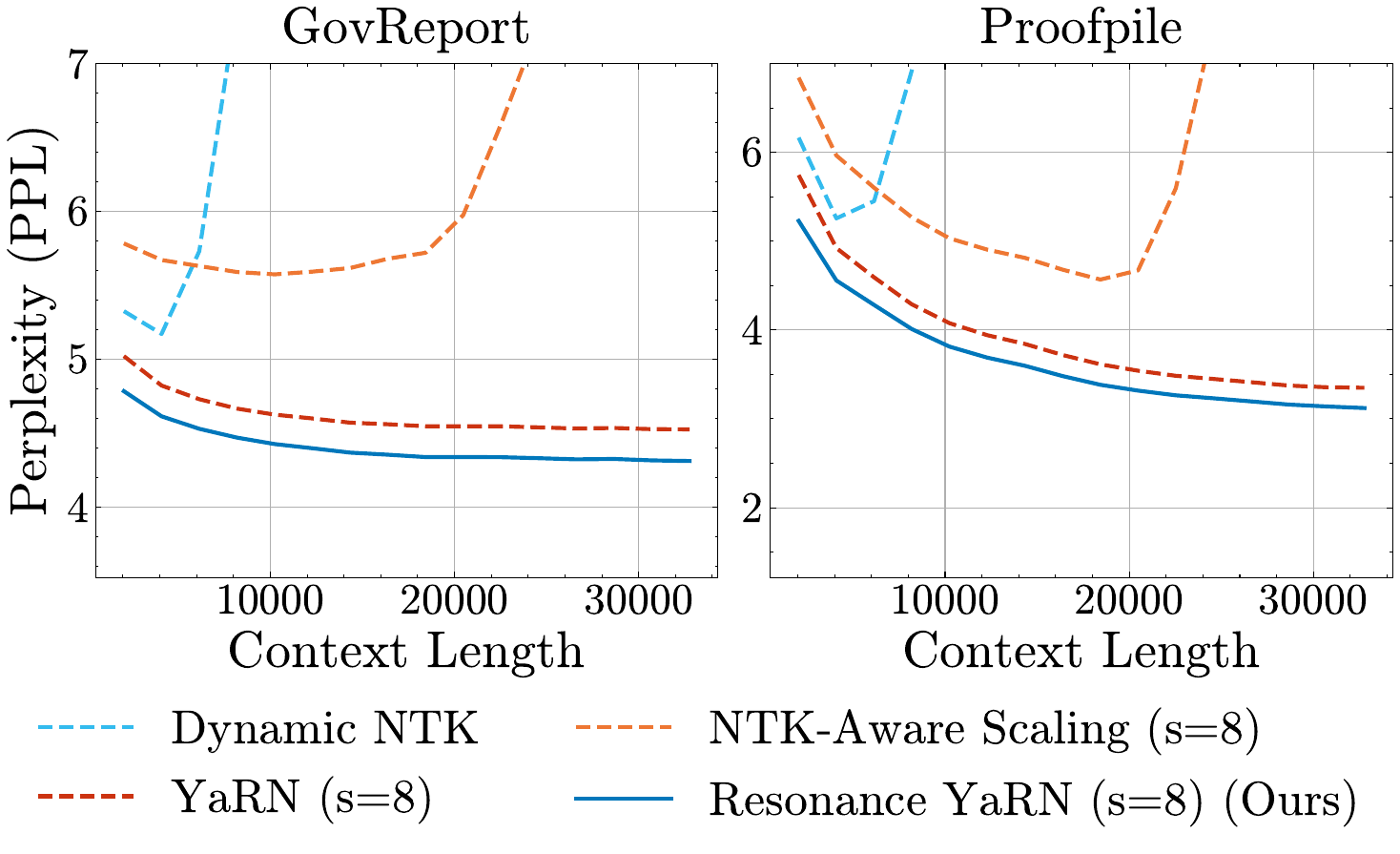}
    \caption{The perplexity of LLaMA-Chat 7B with different position embeddings on GovReport and Proofpile.}
  \label{fig:perplexity}
  \vspace{-5mm}
\end{figure}

We evaluate the model's language modeling performance on GovReport~\citep{huang_efficient_2021} and Proofpile~\citep{azerbayev_zhangir-azerbayevproof-pile_2024}. We randomly select $50$ samples from each dataset and report the final perplexity in text fragments of gradually increased length. We report the results in Figure~\ref{fig:perplexity}. Of the tested methods,~\resyarn~achieves the lowest perplexity across all context lengths. Especially,~\resyarn~achieves a lower perplexity compared to YaRN with the same set of hyperparameters optimized for YaRN, demonstrating the benefit of applying the Resonance technique to existing RoPE scaling methods.

\subsubsection{Real-world Task Evaluation}
\label{sec:real_world_experiments}

Lastly, we test the real-world task performance of LLaMA2-Chat 7B and 13B's performance with different RoPE scaling strategies on L-Eval~\cite{an_l-eval_2023}'s close ended task suite, a long-text LLM benchmark covering a wide range of domains such as school lectures, long conversations and novels. We fine-tune the model with different RoPE scaling strategies using two different strategies: training on shorter sequences (4K length) for more epochs, and training on longer sequences (32K or 16K length) for less epochs. All settings requiring fine-tuning keep the training token count to be approximately 100M. The results are listed in Table~\ref{tab:leval}.

Although no single setting in the experiment achieves the best result on all subtasks, we observe that applying~\resyarn~achieves better average performance in different training settings and model sizes compared to its counterpart YaRN setting. This further proves the compatibility of the Resonance technique and RoPE scaling methods, and the better length extrapolation performance brought by our proposed method.

\section{Conclusion}

We introduce~\methodname, a novel enhancement of RoPE that focuses on minimizing the interpolation of RoPE features for OOD positions, thereby reducing the generalization gap and improving LLM's performance on train-short-test-long (TSTL) scenarios.
Additionally, we present a novel synthetic benchmark,~\datasetname, which provides a fine-grained analysis of the model's TSTL performance regarding various token dependency patterns.
Extensive experiments on our proposed~\datasetname~and two LLM-based evaluations demonstrate~\methodname's efficacy in identifying OOD positions and its compatibility with current RoPE scaling strategies.
Future work includes exploring~\methodname's performance on other foundational models, and the identification of more optimal wavelength combinations for RoPE features.

\section*{Limitations}

Our proposed~\methodname~focus on reducing the interpolation of only RoPE's~\highfreqfeat~on OOD positions. However, this method does not solve the extrapolation issue on RoPE's~\lowfreqfeat, which has been shown to be also detrimental to LLM's length extrapolation performance. Thus, the technique of~\methodname~needs to be combined with another RoPE scaling method that can reduce extrapolation on RoPE's~\lowfreqfeat, e.g., YaRN, to achieve the full potential of LLM in TSTL scenarios. Such combination has been our focus in Section~\ref{sec:llm_experiments}.

Secondly, applying LLMs to long text sequences requires considerations of both performance and efficiency due to the super-linear complexity of Transformers w.r.t input length. As an improvement of the position embeddings, we focus only on improving Transformers' performance in TSTL scenarios. An interesting future direction would be to apply~\methodname~to efficient Transformers for both performance and efficiency enhancements.

Lastly, benchmarking LLMs is still an open question, as there is currently no benchmark to thoroughly test the performance of LLMs, especially on long-sequence tasks. We expect that a more comprehensive long-text benchmark would further improve the validity of the experiment results.

\bibliography{references_new}

\clearpage

\appendix

\section{Proof of Theorem~\ref{thm:generalization}}
\label{appx:proof}
\begin{proof}
   All we need is to prove that for each $\vx \in \mathbb{R}^d$, each $n\in\sN\backslash\{0,\cdots,L-1\}$ and each $i = 0, \dots, 2c-1 $
   we can find  
   $m\in\{0,\cdots,L-1\}$ , such that   $\tilde{f}(\vx,m)_i = \tilde{f}(\vx,n)_i$.
   By definition, it is equivalent to solving the equations:  
$$(\mR^d_{\tilde{\Theta},m}\mW \vx)_i = (\mR^d_{\tilde{\Theta},n}\mW \vx)_i$$
for $m$, given $i$, $n$, and $x$.

   The RoPE feature matrix $\mR^d_{\Theta,m}$ is defined as block-diagonal with $2\times 2$ blocks given by Equation~\ref{eqn:rope-subfeature}. Hence, given $i$, $x$ and $n$, the equation reduces to equality of a linear combination of trigonometric functions:
   $$a \cos{m\tilde\theta_i} + b\sin{m\tilde\theta_i} = a \cos{n\tilde\theta_i} + b\sin{n \tilde\theta_i}$$
   for $a,b \in \mathbb{R}$, depending on $\vx$ and $i$. This equality clearly holds if $m\tilde\theta_i - n\tilde\theta_i$ is a multiple of $2\pi$:
   $$(m-n) \tilde\theta_i = 2\pi k, $$
   for some $k \in \mathbb{Z}$.
   By our construction, $\frac{2\pi}{\tilde\theta_i}$ is a natural number. Hence, to finish the proof that we can solve our initial equation for $m$, we need to show that we can find integer $k$ to satisfy: 
   $$\left(n - \frac{2\pi}{\tilde\theta_i}k \right) \in \{0,\cdots,L-1\}$$
   for $n \in \sN\backslash\{0,\cdots,L-1\}$.
This is where we use the pre-critical dimension condition: for $i=0, \dots, 2c-1$, by definition of $c$, we have the inequality $0 \le \frac{2\pi}{\tilde\theta_i} < L$. Taking $k = \lfloor \frac{n \theta_i}{2\pi} \rfloor$ will give us the required range for $m$ and hence finish the proof.

\end{proof}

\section{Comparison Between Feature Gap and Embedded Vector Distance}
\label{appx:comparison}

Our proposed feature gap metric, as defined in Equation~\ref{eq:gap}, shares similarities with the ``embedded vector distance'' metric introduced by \citet{xiong_effective_2023}:
\begin{equation}
    d(f, \hat f) = \max_{x\in \mathcal X}\min_{{\begin{subarray}{c}
    k\in \{0,\cdots,N-1\}
    \\j\in\{0,\cdots,\hat N -1\}
    \end{subarray}}}\text{dist}[f(x, k), \hat f (x, j)]
\end{equation}
where $\mathcal{X} \subset \mathbb{R}^d$ represents the set of vectors requiring positional embedding. This equation assesses the discrepancy in Rotary Position Embedding (RoPE) before and after a scaling operation. The distance calculation specifically compares the original RoPE, $f(\cdot,\cdot)$, to the scaled RoPE, $\hat{f}(\cdot,\cdot)$, with token positions beginning at zero. It aims to quantify the alterations in position embedding due to the scaling process.

In contrast, our feature gap metric is tailored for a more practical and common scenario, where models are trained or fine-tuned on short sequences using the already scaled RoPE embeddings. This setting emphasizes the generalization gap of the RoPE features between training and testing position ranges. The core hypothesis is that a smaller discrepancy in the RoPE features of new token positions relative to those encountered during training correlates with enhanced model generalization to novel token positions. Our metric diverges from the ``embedded vector distance'' in two significant aspects to better align with our use-case:
\begin{itemize}
\item The distance computation shifts to compare scaled RoPE across different token positions, reflecting the operational context where training involves short sequences (train-short) and testing involves longer sequences (test-long).
\item We modify the token position ranges, $k$ and $j$, to represent token positions observed during training (in-distribution) and testing (out-of-distribution), respectively, to directly measure the generalization gap.
\end{itemize}

This adaptation of the metric allows for a more targeted evaluation of the model's ability to generalize across different token positional distributions, which is critical in scenarios where sequence length varies significantly between training and deployment.

\section{Detailed Experiment Settings}

In this section, we provide the detailed experiment settings for both our synthetic task evaluation on~\datasetname~and LLM-based evaluations on both upstream language modeling evaluation and downstream real-world application evaluations.

\subsection{Synthetic Task Evaluation on~\datasetname}
\label{appx:synthetic_experiments}

For the synthetic task experiments in Section~\ref{sec:synthetic_experiments}, we train a two-layer Transformer on each of the subtasks, with each layer following the configuration of a T5-Small model~\citep{raffel_exploring_2023}. For each subtask, we train the model with different position embeddings on a training set with 10,000 sequence samples of length 64. The validation and test sets each contains 1,000 sequence samples with length 256. The sequences in the training, validation and test sets do not overlap in the first $j+k$ tokens. For all YaRN and~\resyarn~settings, we train the model with YaRN and~\resyarn~applied to the model with a scaling factor $s=4$, which corresponds to the TSTL setting of our evaluation. Each model is trained on each subtask for 150 epochs with a language modeling-style cross-entropy loss. Training was done with AdamW optimizer~\citep{loshchilov_decoupled_2019}, using learning rate $2\times 10^{-4}$ and weight decay $1\times 10^{-2}$. We use a batch size of $128$ for all experiments. All hyperparameters were tuned to maximize YaRN's validation set performance on the Semi-Recurrent subtask. All synthetic task evaluations were performed on a single NVIDIA V100 32G GPU.

\subsection{LLM Evaluations}
\label{appx:llm_setting}
For the LLM-based evaluations in Section~\ref{sec:llm_experiments}, we fine-tune LLaMA2-Chat 7B or LLaMA2-Chat 13B~\citep{touvron_llama_2023-1} after replacing its original RoPE position embedding with RoPE scaled with different strategies:
\begin{itemize}
    \item \textbf{NTK-Aware Scaling}~\citep{bloc97_ntk-aware_2023,xiong_effective_2023,liu_scaling_2023}, which scales the base $b$ in Equation~\ref{eqn:theta} to $s\cdot b$, where $s$ is the scaling factor. We evaluate the performance without fine-tuning as used in~\citet{bloc97_ntk-aware_2023}.
    \item \textbf{Dynamic NTK-Aware Scaling}~\citep{peng_yarn_2023,roziere_code_2023}. This method dynamically computes the scaling factor considering the current sequence length $L_c$ and the original context window length $L$: $s=\frac{L_c}{L}$. Due to the high cost of frequently recomputing RoPE features, we evaluated its performance without fine-tuning.
    \item \textbf{YaRN}~\citep{peng_yarn_2023}. We evaluate its performance after fine-tuning.
\end{itemize}

For NTK-Aware scaling and Dynamic NTK-Aware scaling settings, we replace the original RoPE position embeddings in the model with the scaled ones and test their performance without fine-tuning following~\citep{bloc97_ntk-aware_2023,peng_yarn_2023}. For YaRN and~\resyarn~settings, we fine-tune the model for approximately 100M tokens on PG19's training set~\citep{rae_compressive_2019}. Our target scaled length for the 7B and 13B models is 32K and 16K, respectively, which corresponds to a scaling factor $s=8$ and $s=4$ for the position embeddings of the two models.

For both the long-sequence perplexity evaluation in Section~\ref{sec:perplexity_experiments} and real-world task evaluations in Section~\ref{sec:real_world_experiments}, the hyperparameters for the LLM experiments follow the configurations provided in~\citet{peng_yarn_2023}\footnote{\url{https://github.com/jquesnelle/yarn}.}, with the only modification that we fine-tune the model on approximately 100M tokens. More specifically, we use $\alpha=1$ and $\beta=32$ for YaRN and~\resyarn as suggested by~\citet{peng_yarn_2023}. The model was trained with a language modeling-style cross entropy loss. Training was done with the AdamW optimizer~\citep{loshchilov_decoupled_2019} using learning rate $2\times 10^{-5}$ and weight decay $1\times 10^-2$. We use a batch size of $1$ on each of the GPUs. The learning rate warm-up is applied to the first $5\%$ of the total training steps. Models were fine-tuned with BF16 precision, FlashAttention 2~\citep{DBLP:journals/corr/abs-2307-08691} and DeepSpeed ZeRO-3 Offload~\citep{DBLP:conf/usenix/0015RARYZ0H21} on four NVIDIA A100 40G GPUs.

For the real-world task evaluations in Section~\ref{sec:real_world_experiments}, we further compare two different fine-tuning strategies:
\begin{enumerate}
    \item \textbf{Fine-tuning on long sequences for less epochs.} We directly fine-tune the model on the target sequence lengths after applying the scaled position embeddings. For LLaMA2-Chat 7B and 13B, we fine-tune the model on sequences with length 32,768 for 50 steps and sequences with length 16,384 for 100 steps, respectively.
    \item \textbf{Finetuning on short sequences for more epochs.} We fine-tune the model on the original pre-training sequence length after applying the scaled position embeddings. For both LLaMA2-Chat 7B and 13B, we fine-tune the model on sequences with length 4,096 for 400 steps.
\end{enumerate}

\end{document}

%% file: math_commands.tex
\def\eqref#1{equation~\ref{#1}}
\def\floor#1{\lfloor #1 \rfloor}
\def\1{\bm{1}}

\def\vk{{\bm{k}}}

\def\vq{{\bm{q}}}

\def\vx{{\bm{x}}}

\def\mR{{\bm{R}}}

\def\mW{{\bm{W}}}

\DeclareMathAlphabet{\mathsfit}{\encodingdefault}{\sfdefault}{m}{sl}
\SetMathAlphabet{\mathsfit}{bold}{\encodingdefault}{\sfdefault}{bx}{n}

\def\sN{{\mathbb{N}}}

\def\sX{{\mathbb{X}}}

\newcommand{\R}{\mathbb{R}}

\usepackage{cancel}